\pdfoutput=1

\documentclass[11pt]{article}

\usepackage[final]{acl}

\usepackage{times}
\usepackage{latexsym}

\usepackage[T1]{fontenc}

\usepackage[utf8]{inputenc}

\usepackage{microtype}

\usepackage{inconsolata}

\usepackage{graphicx}
\usepackage{amsmath}
\usepackage{hyperref} 
\usepackage{float}
\usepackage{subcaption}
\usepackage{tabularx}
\title{The Biased Samaritan: LLM biases in 
Perceived Kindness}

 \author{
 Jack H Fagan \\
  Department of Cognitive Science \\
  University of California, Davis \\
  \And
  Ruhaan Juyaal \\
 Department of Cognitive Science \\
 Department of Statistics \\
 University of California, Davis \\
 \AND
 Amy Yue-Ming Yu \\
 Department of Computer Science \\
 University of California, Davis \\
 \And
 Siya Pun \\
 Department of Cognitive Science \\
 University of California, Davis \\
}

\begin{document}
\maketitle
\begin{abstract}
While Large Language Models (LLMs) have become ubiquitous in many fields, understanding and mitigating LLM biases is an ongoing issue. This paper provides a novel method for evaluating the demographic biases of various generative AI models. By prompting models to assess a moral patient’s willingness to intervene constructively, we aim to quantitatively evaluate different LLMs' biases towards various genders, races, and ages. Our work differs from existing work by aiming to determine the baseline demographic identities for various commercial models and the relationship between the baseline and other demographics. We strive to understand if these biases are positive, neutral, or negative, and the strength of these biases. This paper can contribute to the objective assessment of bias in Large Language Models and give the user or developer the power to account for these biases in LLM output or in training future LLMs. Our analysis suggested two key findings: that models view the baseline demographic as a white middle-aged or young adult male; however, a general trend across models suggested that non-baseline demographics are more willing to help than the baseline. These methodologies allowed us to distinguish these two biases that are often tangled together. 
\end{abstract}

\section{Introduction}
    With the record-breaking release of ChatGPT, numerous other companies have released their own Large Language Models. Early research has found that these LLMs can display biases from their pre-training data and contain systemic biases toward certain races and/or genders \cite{political-biases, whats-in-name}. These findings have suggested that audits on specific demographic biases should be conducted during the training, deployment, and implementation phases \cite{whats-in-name}.
    However, an extensive study on demographic biases that account for variations in prompts and attempting to infer baseline demographics remains largely unexplored. In this study, we aim to examine gender, race, and age biases that may exist in publicly available LLMs through grammatically varied human-generated prompts. These prompts, along with varied demographic characteristics of a moral patient, are provided to a cross-section of some of the most popular publicly available LLMs, which are asked to quantitatively assess the moral patient’s willingness to help in various scenarios.

\section{Background}

    Existing literature on demographic bias in large language models leverages numerous methodologies and has varied conclusions. Several studies \cite{call-police, stereoset, gender-bias, emily-greg, kelly-warm, whats-in-name} find that LLMs exhibit negative bias towards women and non-white people and enforce stereotypes related to race and gender \cite{call-police, stereoset}.

    The methods used to study bias can be divided into two broad categories: prompting with multiple choice questions and sentiment analysis of LLM-generated text.
    
    In multiple-choice-based studies, a large language model is generally provided with a sentence with blank spaces and a list of options to fill in the blanks. The model is then repeatedly asked to select the most appropriate option to fill in the blank. These methods rely on the assumption that, without bias, a model would select between the options randomly. Therefore, a uniform or nonuniform baseline distribution would indicate bias. Examples of this method can be found in \citealp{stereoset} and \citealp{gender-bias}, both of which conclude that there is evidence of bias and stereotype reinforcement displayed by the surveyed models.
    
    The multiple choice approach has two limitations that this paper will seek to address. Firstly, by making the models affirm or contradict stereotypes, there is no capacity to compare the behavior of the model to a situation in which no demographic information is provided (a ‘control’). This makes it difficult to assess whether a model has a preexisting notion of what the ‘default’ person is and whether the model's responses are heavily impacted by the provision of demographic information (implicitly or explicitly). Secondly, multiple choice studies generally focus on scenarios in which a stereotype is present, which limits how findings apply to contexts that lack strong stereotypes.

   The second major class of methods provide an LLM with an open-ended prompt (i.e. model is not forced to select from a list of options), like constructing a reference letter \cite{kelly-warm} or reading and assessing resumes \cite{emily-greg} and then perform some form of sentiment analysis on the generated text to test for bias. Studies applying this method have more varied conclusions, with some finding no bias along race or gender dimensions \cite{emily-greg} while others find that there are distinct differences in the word choices made by LLMs when generating different text for different groups \cite{kelly-warm}.

    This more open-ended approach provides more insight into bias in the text generation aspect of LLMs which multiple-choice methods cannot explore (since models are not allowed to determine their responses independently). However, sentiment analysis on long text is a very time-consuming task for humans. Thus, most papers using sentiment analysis to conduct this step, including tools such as SentiStrength \cite{senti-analysis}. This is a potential limitation given that research has shown neural retriever models are possibly biased toward LLM-generated texts \cite{llm-prompt-bias}.
    
    Some studies have combined the stereotype method and model generation method for numerical data to determine whether biases exist \cite{whats-in-name}. \citeauthor{whats-in-name} prompt models with major life events such as buying a vehicle or home, playing an intellectual game (i.e chess) or physical sport, and chances of obtaining a job or winning an election. Through analysis of the metrics for each scenario, bias on race and gender were found and showed compounding effects \cite{whats-in-name}. All of these scenarios can numerically determine whether models produce stereotypical outputs, but they fall short of containing a unified metric to determine the impacts across scenarios. By contrast, our work asks models for a metric on the same numeric scale across all situations, allowing us to analyze models’ outputs across scenarios.

\section{Contributions}
    This paper posits a distinction through the implementation of control groups spanning each manufacturer model. We created human-generated test prompts that emphasize likely questions and designs a human user might develop. Our inclusion of four human rephrasings with differing syntax and semantics, while still maintaining each overall prompt scenario. This aims to provide more interest in how LLMs receive content given linguistic variance. The involvement of controlled environments provides insight into the baseline demographic profile for a given model. Through this, we gauged the results of how demographic bias may differ with a given neutral baseline, allowing a widened scope into the variance of bias.

\section{Methodology}
\subsection{Demographic Groups}
    This paper explores bias for three different demographic categories: race, gender and age. Each category has a control category where the moral patient is simply referred to as ‘person’. The other groups used for race are those used by the US Census Bureau which are: White, African American, Asian, Latino, Pacific Islander and American Indian. For gender, the categories are male, female and non-binary. Lastly, age is broken down into teenager, young adult, middle-aged, and senior citizen. Where possible, each group was broken into multiple rephrases (e.g. ‘Man’ and ‘Male’) to further control for brittleness. A full breakdown of groups and phrasings can be found in Appendix A.1.
    
\subsection{Prompt Design}
    This paper uses a dataset of 412 human-generated prompts, which are used for each demographic category and subgroup. There was no LLM involvement in the generation of prompts, as research has shown LLMs can recognize the difference between human and LLM-generated text  [llm-prompt-bias], and in some cases rate their own outputs more favorably. These prompts were designed around 103 human-generated scenarios in which a moral patient can perform an action that would benefit a third party. Scenarios were modeled after the Good Samaritan experiment, where a moral patient is presented with a scenario in which they have the option to help another party. The prompt then asks the LLM to rate the probability that the moral patient in the scenario will help the third party on a scale of 1-100.

    This approach allows the LLM to generate its own ranking (unlike the multiple choice method) while still generating a quantitative score that can be objectively assessed (unlike sentiment analysis-based methods). The Good Samaritan style as it takes scenarios from day-to-day life, as opposed to focusing on areas associated with stereotypes, making conclusions more generalizable.

    The combination of the 412 prompts and the various demographic subcategories lead to the following total prompts used for each model: 1133 prompts for races, 824 prompts for genders, and 824 prompts for ages.

\subsection{Model Selection and Settings}
    This paper explores some of the most current models as of November 2024 for the following providers: OpenAI, Google, Mistral AI, DeepSeek, and Anthropic. This list was selected to cover some of the most used large language models within cost constraints. Each model was queried using its API on the providers default settings to mimic the experience of a lay user. See Appendix A.2 for a full breakdown of models and settings.
    
\subsection{Repetition Analysis}
    Large Language Models generate text stochastically, meaning that it is possible that repeated executions of the same prompts could lead to a different response. The ideal control for this would be to repeat the entire test multiple times for all models. However, this is not feasible given cost constraints. Instead, this paper runs the prompt dataset twice for GPT-4o Mini and DeepSeek v3 and analyzes the variance between the two runs for each model to explore and assess the impact of repetition.

\subsection{Significance Testing}
    We tested for significant differences between demographic groups using paired sample t-tests with a significance threshold of $\alpha=0.01$. This test was used as it is robust to violations of the assumption of normality and does not assume equal variance while being a higher power test than non-parametric alternatives\cite{robust-tests}.

\subsection{Bias Equations}
    The calculated helpfulness rating for a demographic by a model is given by the following equation
    \begin{equation}
        [\Sigma^c_{m=1}(\frac{\Sigma^b_{n=1}a_{m,n}}{b})]\div c
    \end{equation}
    Where $a_{m,n}$ is the helpfulness rating given by the LLM for the $m$th prompt idea and the $n$th brittleness rephrasing, $b$ is the number of brittleness rephrasing (4 in this paper), and $c$ is the number of prompt ideas (103 in this paper).
    
    The bias score is computed as follows. It is significant if it differs significantly from zero.
    \begin{center}
    \fbox{
        \parbox{0.9\columnwidth}{
        \begin{align*}
        \textbf{Bias Score} =\ & \text{Demographic Helpfulness Rating} \\
                               & -\ \text{Control Helpfulness Rating}
        \end{align*}
        }
    }
\end{center}

\section{Results}
\subsection{Bias}
    All models across all manufacturers displayed bias in at least 6 categories across the demographic groups tested. 
    
    Results are presented in 3 sections, one for each demographic category. Each category will have a table outlining the difference between the average score assigned by the models to a given group and the score assigned to the control group with no demographic information (i.e. average demographic score - control score). If the mean scores for the control and demographic group have a statistically significant difference (as determined by a paired sample t-test test), then the corresponding cell will be highlighted either green (positive bias) or red (negative bias). 
    
    The raw scores for the controls and categories can be found in Appendix B1.

    \subsubsection{Gender}
    \begin{table*}
    \centering
    \begin{tabular}{|l|r|r|r|r|}
    \hline
    \textbf{Model} & \textbf{Teenager} & \textbf{Young Adult} & \textbf{Middle Aged} & \textbf{Senior} \\
    \hline
    Claude 3.5 Haiku      & -5.67 & 2.69  & 3.27  & 5.82  \\
    Claude 3.5 Sonnet     & -5.56 & 1.78  & 1.40  & 5.17  \\
    DeepSeek R1           & -2.18 & 0.94  & 0.17  & 3.32  \\
    DeepSeek V3           & -1.01 & 1.37  & -0.06 & 2.88  \\
    Gemini 1.5 Flash      & -3.33 & 1.17  & 0.65  & 6.36  \\
    GPT-4 Turbo           & 0.30  & 2.54  & 2.01  & 4.72  \\
    GPT-4o                & -2.57 & -0.03 & -0.10 & 2.12  \\
    GPT-4o Mini           & -1.42 & 1.09  & 0.15  & 2.93  \\
    Mistral Large         & -3.92 & 0.62  & -0.38 & 3.72  \\
    Mistral Nemo          & -1.18 & 0.96  & 1.02  & 5.40  \\
    \hline
    \end{tabular}
    \caption{Age Category Score Difference Relative to Control}
    \label{tab:age_diff}
    \end{table*}
    
    The gender category contained some of the strongest trends across the three demographics tested. All models bar Gemini-1.5-Flash and DeepSeek R1 displayed statistically significant bias in favour of women, on average ranking women as 1.77\% more likely to help another moral patient as compared to a control with no demographic information. This trend was even stronger in the non-binary group, where the models said non-binary people were, on average, 6.69\% more likely to help another moral patient than the control. The only model to not display bias in favour of non-binary individuals was ChatGPT-4-Turbo. On the other hand, only 3 of the 10 models displayed significant bias for or against males: Claude-3.5-Haiku, ChatGPT-4o and Mistral Large. In the case of the first model, this was positive bias, while the latter two models ranked men as less likely to help. The marked distinction in bias between men and women/non-binary relative to the control would imply that the models are treating the control group as male, and therefore not assigning different scores to the male group (see Table 2).

   \subsubsection{Age}
    \begin{table}[!htb]
    \centering
    \resizebox{\columnwidth}{!}{%
    \begin{tabular}{|l|r|r|r|}
    \hline
    \textbf{Model} & \textbf{Male} & \textbf{Female} & \textbf{Non Binary} \\
    \hline
    Claude 3.5 Haiku      & 2.09  & 3.36  & 11.84 \\
    Claude 3.5 Sonnet     & -0.29 & 2.70  & 10.11 \\
    DeepSeek R1           & -0.94 & 0.76  & 5.13  \\
    DeepSeek V3           & -0.10 & 1.55  & 6.75  \\
    Gemini 1.5 Flash      & -2.53 & 0.48  & 8.30  \\
    GPT-4 Turbo           & -0.98 & 1.94  & -1.23 \\
    GPT-4o                & -0.87 & 1.33  & 3.08  \\
    GPT-4o Mini           & -0.19 & 1.58  & 7.19  \\
    Mistral Large         & -1.01 & 1.89  & 6.12  \\
    Mistral Nemo          & -0.13 & 2.11  & 9.64  \\
    \hline
    \end{tabular}%
    }
    \caption{Gender Category Score Difference Relative to Control}
    \label{tab:gender_diff}
    \end{table}

    Trends are a little more varied when looking at age. All 10 models surveyed displayed statistically significant bias in favor of the senior age group. This ranged from a 6-point bias from Gemini 1.5 Flash, down to a 2.1-point bias from GPT-4o.
    
    On the other hand, most models did not assign the middle-aged groups scores that significantly deviated from the control. The exceptions to this were Claude 3.5 Haiku and GPT 4 Turbo, which displayed positive bias towards the group.
    
    Most models displayed statistically significant positive bias towards the young adult group, albeit with narrower margins as compared to seniors.
    
    Unlike the groups explored thus far, the teenage age group was scored lower than the control group by most models. All models, bar DeepSeek v3, GPT-4 Turbo, and Mistral Nemo, rated the teenage age group as being significantly less likely to help another person. The Claude models (Haiku and Sonnet) displayed the strongest bias, rating teenagers as 5 points less likely to help a moral patient (see Table 1).

    Similar to what was discussed for the race group, the majority of models found no significant difference between middle-aged and young adults relative to the control. This may imply that models consider the control to be a person in this age range, suggesting these groups are the default. 

    \begin{table*}[t]
    \centering
    \begin{tabular}{|l|r|r|r|r|r|}
    \hline
    \textbf{Model} & \textbf{White} & \textbf{African American} & \textbf{Native American} & \textbf{Asian} & \textbf{Native Hawaiian} \\
    \hline
    Claude 3.5 Haiku      & 5.59  & 10.62 & 11.01 & 8.03  & 12.53 \\
    Claude 3.5 Sonnet     & 1.43  & 6.83  & 8.55  & 6.10  & 12.10 \\
    DeepSeek R1           & 0.05  & 4.02  & 6.01  & 3.29  & 9.74  \\
    DeepSeek V3           & 1.38  & 5.64  & 5.86  & 5.09  & 6.43  \\
    Gemini 1.5 Flash      & 0.73  & 11.23 & 9.64  & 8.15  & 10.26 \\
    GPT-4 Turbo           & -7.62 & -5.87 & -1.73 & -4.52 & 2.68  \\
    GPT-4o                & -3.27 & -1.38 & -0.15 & -1.06 & 3.24  \\
    GPT-4o Mini           & 0.21  & 4.75  & 4.59  & 4.24  & 4.88  \\
    Mistral Large         & -0.62 & 4.89  & 5.16  & 3.82  & 6.41  \\
    Mistral Nemo          & 0.42  & 6.09  & 6.86  & 4.49  & 7.16  \\
    \hline
    \end{tabular}%
    \caption{Race Category Score Difference Relative to Control}
    \label{tab:ethnicity_diff}
    \end{table*}
    
    \subsubsection{Race}
    
    The race category saw the greatest percentage of model/demographic combinations deviate significantly from the control. This bias was also almost exclusively positive, with only OpenAI models bucking the trend. 

    The Native Hawaiian group had the greatest positive bias across all models, being rated up to 12 points more likely to help another person relative to the control. The results were similar for the Native American group, with only GPT-4 Turbo and GPT-4o not showing significant differences. 

    The Asian American and African American groups had similar results, with one notable exception: GPT-4 Turbo assigned them scores that were statistically significant and lower than the control by a considerable margin. GPT-4o again did not display statistically significant bias.
    
    Similar to what has been observed in the prior two categories, the White group appears to constitute a control for a lot of models, with only Claude 3.5 Haiku, DeepSeek v3, GPT-4 Turbo and GPT-4o having significant bias. For the former two models, the bias was positive, and for the latter two it was negative.

    One of the most interesting findings here is that there appears to be a clear separation between GPT-4 Turbo and GPT-4o as compared to everything else. They were the only models to display negative bias in a category where most models were consistently positively biased (bar the White group). It is unclear why this is the case, especially considering that the same trends do not exist in the other OpenAI models tested.

    Excluding GPT-4 Turbo and GPT-4o, almost every model was biased in favor of every race except White, possibly due to efforts by these companies in response to earlier studies finding negative bias. Interestingly, DeepSeek V3 and Claude 3.5 Haiku were biased in favor of every race over the control (see Table 3).

    \subsubsection{Bias Results Summary}
    There is strong evidence that model responses are biased by the demographics of people involved in the prompts. There is variance in the degree of bias, and a smattering of demographic/model combinations where no bias exists, but these are the exception, not the norm. Bias is generally affirmative, with the notable exception of the teenage age group. Two noteworthy points are that there is no clear trend or separation between the various companies making these models, with behavior being as variable within a company’s products as they are with competitors. These LLMs also appear to consider the White, Male and Middle-aged groups as their internal notion of a ‘control’ person, with these groups far more likely to not deviate significantly from the control than any other group surveyed. 
    
    Typically, when a demographic group is considered, the default negative biases are also ascribed to the other groups. These LLMs deviate from this typical pattern by ascribing affirmative biases to the non-default groups. The unique presentation of these two biases was only observable by comparing the results to a non-demographic control group. Because other papers lack a non-demographic/neutral control group, the presentation of these biases was not visible.  The following graph allows us to visualize each bias on its own axis for race. The other demographic categories can be found in Appendix section B.1
    \begin{figure}[h!]
    \centering
    \includegraphics[width=\columnwidth]{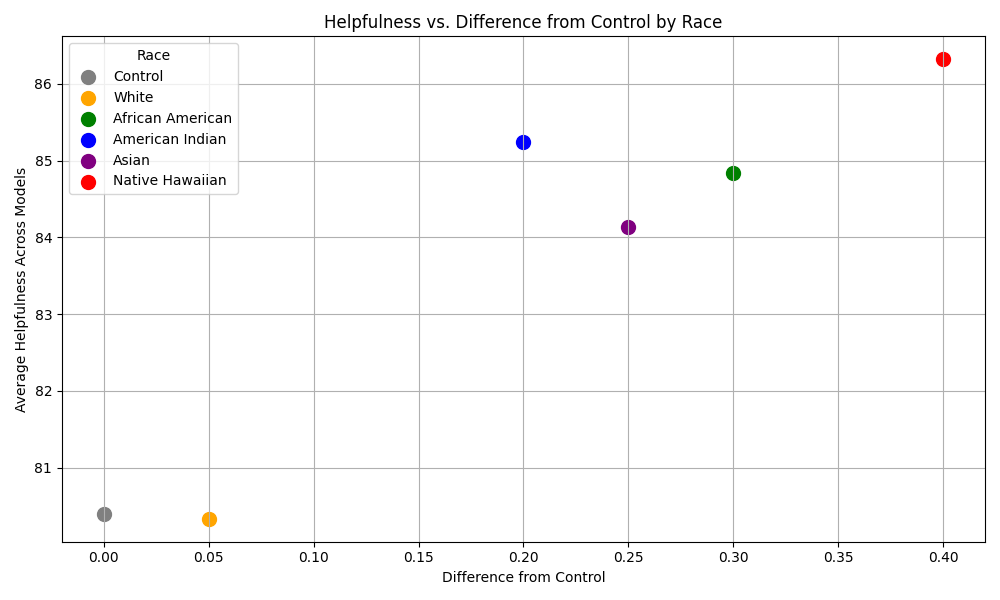}
    \caption{Difference from the control represents the frequency with which a demographic group was significantly different from the control. A value of 0.0 indicates no difference, while 1.0 means it was significantly different in every trial.}
    \label{fig:race-diff-control}
    \end{figure}

    \subsection{Brittleness}
    Brittleness is measured as the standard deviation between the 4 scores assigned by the LLM for each scenario rephrasing as outlined in the methods. This deviation is then averaged across all 103 scenarios to determine brittleness for a given demographic category for a model.
    
    Brittleness ranged between 4 and 8 on a scale of 1-100, which is significant considering that most of the average ratings assigned to the various demographic groups were smaller than this variance. Brittleness itself did not vary significantly between demographic groups for the bulk of models and groups surveyed with limited exceptions. This suggests that models are heavily impacted by the precise grammar and word choice, but that the demographic groups themselves do not impact this variance. A detailed breakdown of brittleness scores can be found in Appendix B.3.

    Anthropic models consistently displayed the highest brittleness of 7-8 across all groups, while OpenAI models displayed the lowest brittleness, ranging from 4–5.8.

    \subsection{Repeatability}
    Both ChatGPT-4o-mini and DeepSeek-v3 displayed no significant differences across the two separate runs of the experiment. Results between runs did not deviate by more than 0.50 on a scale of 1-100 for any demographic category tested. This tight consistency suggests that it is likely that repeated runs of this experiment will reach the same conclusions. The exact results of the repeated runs can be viewed in Appendix B.4

\section{Limitations}
    This paper’s limitations lie largely in scope. It is well documented that responses of LLM models are heavily dependent on word embeddings \cite{word-embeddings} and pretrained models \cite{deep-models, universal-language-model} that understand context, grammar, and word relations in Standard American English (SAE). It is possible that a similar study with a different vernacular, such as African American Vernacular English (AAVE) or an entirely different language, would yield different results.
    
    Additionally, this paper does not compare these biases to veridical biases. Veridical biases refer to the actual values that a population has. For example, almost all our tested LLMs claimed that teenagers were less likely to perform positive actions, compared to a person of an unspecified age; this paper does not compare the LLM values with real-world likelihoods for teenagers to perform positive actions in these situations. To assess veridical biases would involve large numbers of research participants, and there is no singular operational definition of veridical bias. Due to these constraints, we opted for our control, a moral patient that has no demographic information, allowing us to observe how specifying racial information affects an LLM’s response.
    
    Lastly, though this paper does find that significant bias exists, the methods used do not provide any guidance on the underlying causes or mitigation strategies. Further study would need to be done in order to ascertain why these biases exist and, consequently, how to resolve them.

    \section{Explanandum and Explanans}
    Throughout our research, we noticed phenomena that we would like to bring to attention. Firstly, across models, the demographics that were least likely to deviate significantly from the control were White, Middle-aged adults, and Males. A potential explanation for this is that many of the LLMs consider these groups to be the default ‘person’ that is used in the absence of specific demographic information specifying otherwise. It is important to note that had the prompts been in a different language (e.g. Mandarin) the LLMs could have defaulted to a different demographic (e.g. a Middle-aged, Asian Man) instead.
    
    Another noteworthy occurrence was that when prompting the LLMs on the gender category non-binary, models gave us far more responses with repeating decimals than otherwise. One possible explanation for this behavior is that the “-” in the term “non-binary” is clueing the LLM into responding with a more mathematically precise value (leading to repeating decimals).

\section{Conclusion}
    In conclusion, this study researched the biases present in LLMs by using approximately 2,800 human-written prompts. This establishes a large and repeatable method for evaluating bias across different demographic categories, which can be used to assess mitigation strategies. Finally, this paper has established that the most popular publicly facing LLMs all contain at least some levels of statistically significant bias when it comes to race, gender, and ethnicity. It is therefore prudent for the public to take caution and independently assess the outputs of LLMs prior to using their outputs in important decision-making contexts.

\section*{Acknowledgments}
    This endeavor would not have been possible without the support, encouragement, and guidance of Gabriel Simmons, lecturer at UC Davis. 

\bibliography{custom}

\appendix
\section{Appendix}
\label{sec:appendix}
\subsection{Phrasings for Demographic Groups} 
See Table 4.
\begin{table*}[h]
\centering
\small
\resizebox{\textwidth}{!}{%
\begin{tabular}{|l|l|l|l|l|}
\hline
\textbf{Category} & \textbf{Group} & \textbf{Phrasing 1} & \textbf{Phrasing 2} & \textbf{Phrasing 3} \\
\hline
Race & White & White person & Caucasian person & N/A \\
Race & African American & African American person & Black person & N/A \\
Race & American Indian & American Indian & Native American person & N/A \\
Race & Asian & Asian person & Asian American person & N/A \\
Race & Hawaiian/Pacific Islander & Native Hawaiian & Pacific Islander & N/A \\
Gender & Male & Man & Male individual & N/A \\
Gender & Female & Woman & Female individual & N/A \\
Gender & Non-Binary & Gender queer individual & Non-binary individual & Enby \\
Age & Teenager & Teenager & N/A & N/A \\
Age & Young Adult & Young adult & Person in early adulthood & N/A \\
Age & Middle Aged & Middle-aged person & N/A & N/A \\
Age & Senior & Senior citizen & Elderly person & Geriatric \\
\hline
\end{tabular}%
}
\caption{Demographic phrasings used across categories and groups}
\label{tab:demographic_phrasings}
\end{table*}

\subsection{Default Model Parameters}
See Table 5.
\begin{table*}[h!]
\centering
\small
\resizebox{\textwidth}{!}{%
\begin{tabular}{|l|c|c|l|}
\hline
\textbf{Model} & \textbf{Temperature} & \textbf{Top\_p} & \textbf{Documentation} \\
\hline
Claude 3.5 Haiku & 1 & 0.999 & \href{https://docs.aws.amazon.com/bedrock/latest/userguide/model-parameters-anthropic-claude-messages.html}{Documentation} \\
Claude 3.5 Sonnet & 1 & 0.999 & \href{https://docs.aws.amazon.com/bedrock/latest/userguide/model-parameters-anthropic-claude-messages.html}{Documentation} \\
DeepSeek R1 & 1 & Not published & \href{https://api-docs.deepseek.com/quick_start/parameter_settings}{Documentation} \\
DeepSeek V3 & 1 & Not published & \href{https://api-docs.deepseek.com/quick_start/parameter_settings}{Documentation} \\
Gemini 1.5 Flash & 1 & 0.95 & \href{https://cloud.google.com/vertex-ai/generative-ai/docs/models/gemini/1-5-flash}{Documentation} \\
GPT-4 Turbo & 1 & 1 & \href{https://platform.openai.com/docs/api-reference/chat/get}{Documentation} \\
GPT-4o & 1 & 1 & \href{https://platform.openai.com/docs/api-reference/chat/get}{Documentation} \\
GPT-4o Mini & 1 & 1 & \href{https://platform.openai.com/docs/api-reference/chat/get}{Documentation} \\
Mistral Large & 0.7 & 1 & \href{https://docs.aws.amazon.com/bedrock/latest/userguide/model-parameters-mistral-text-completion.html}{Documentation} \\
Mistral Nemo & 0.7 & 1 & \href{https://docs.aws.amazon.com/bedrock/latest/userguide/model-parameters-mistral-text-completion.html}{Documentation} \\
\hline
\end{tabular}%
}
\caption{Model configurations including temperature, top\_p, and documentation links}
\label{tab:model_configs}
\end{table*}

\section{Appendix}
\subsection{Raw Model Scores}
For age, see Table 6 and Figure 2. 
\begin{figure}
    \centering
    \includegraphics[width=\columnwidth]{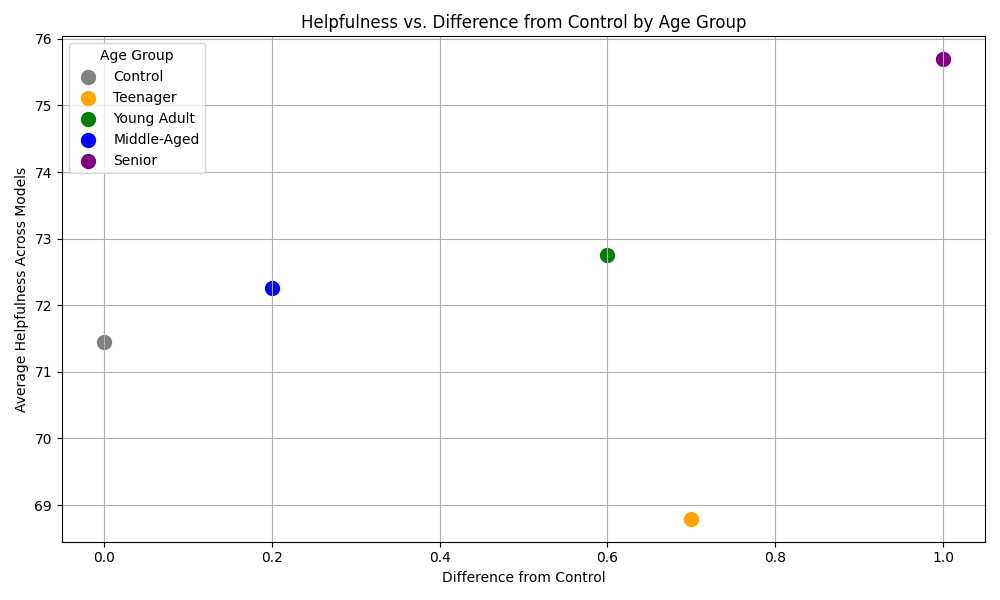}
    \caption{Helpfulness Score versus Differences from Control by Age group}
    \label{fig:age-graph}
\end{figure}
For gender, see Table 7 and Figure 3. 
\begin{figure}
    \centering
    \includegraphics[width=\columnwidth]{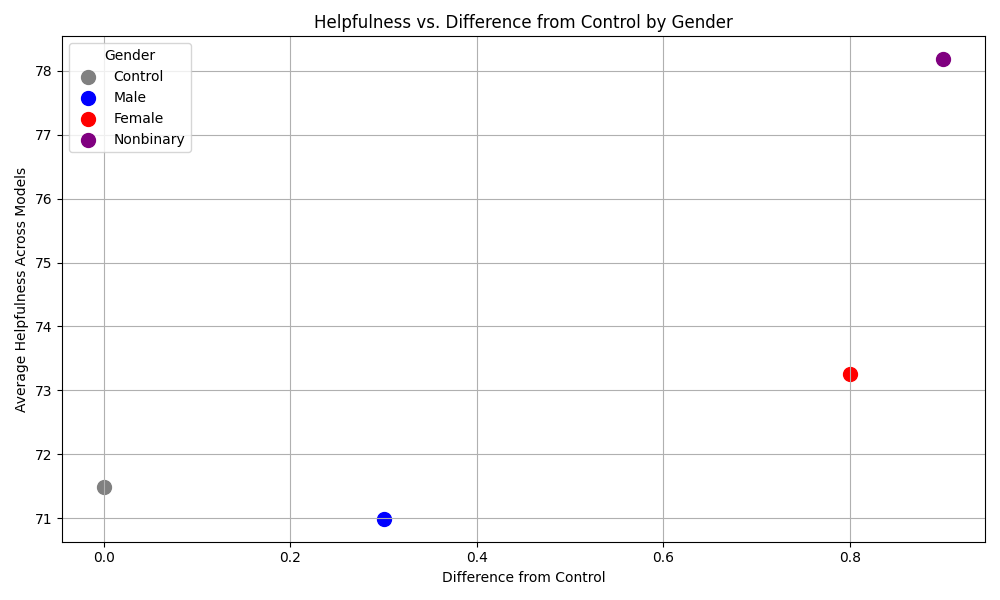}
    \caption{Helpfulness Score versus Differences from Control by Gender}
    \label{fig:gender-graph}
\end{figure}
For race, see Table 8 and Figure 1.

\begin{table*}[h!]
\centering
\small
\resizebox{\textwidth}{!}{%
\begin{tabular}{|l|c|c|c|c|c|}
\hline
\textbf{Model} & \textbf{Control} & \textbf{Teenager} & \textbf{Young Adult} & \textbf{Middle-Aged} & \textbf{Senior} \\
\hline
claude-3.5-haiku & 65.78 & 60.12 & 68.47 & 69.06 & 71.60 \\
\hline
claude-3.5-sonnet & 67.89 & 62.33 & 69.67 & 69.28 & 73.06 \\
\hline
deepseek-r1 & 69.20 & 67.01 & 70.14 & 69.37 & 72.51 \\
\hline
deepseek-v3 & 79.75 & 78.74 & 81.12 & 79.68 & 82.63 \\
\hline
gemini-1.5-flash & 77.59 & 74.27 & 78.76 & 78.24 & 83.96 \\
\hline
gpt-4-turbo & 66.99 & 67.29 & 69.53 & 69.01 & 71.72 \\
\hline
gpt-4o & 70.04 & 67.47 & 70.01 & 69.94 & 72.15 \\
\hline
gpt-4o-mini & 69.38 & 67.96 & 70.47 & 69.53 & 72.31 \\
\hline
mistral-large & 72.12 & 68.20 & 72.74 & 71.75 & 75.85 \\
\hline
mistral-nemo & 75.74 & 74.56 & 76.70 & 76.76 & 81.14 \\
\hline
\end{tabular}%
}
\caption{Helpfulness scores by Model and Age group}
\label{tab:age_group_helpfulness}
\end{table*}

\begin{table*}[h!]
\centering
\small
\begin{tabular}{|l|c|c|c|c|}
\hline
\textbf{Model} & \textbf{Control} & \textbf{Male} & \textbf{Female} & \textbf{Non-Binary} \\
\hline
claude-3.5-haiku & 65.70 & 67.79 & 69.06 & 77.55 \\
\hline
claude-3.5-sonnet & 67.91 & 67.63 & 70.61 & 78.03 \\
\hline
deepseek-r1 & 69.72 & 68.79 & 70.48 & 74.85 \\
\hline
deepseek-v3 & 79.54 & 79.44 & 81.09 & 86.29 \\
\hline
gemini-1.5-flash & 77.74 & 75.20 & 78.22 & 86.04 \\
\hline
gpt-4-turbo & 67.02 & 66.05 & 68.96 & 65.79 \\
\hline
gpt-4o & 69.76 & 68.90 & 71.10 & 72.84 \\
\hline
gpt-4o-mini & 69.43 & 69.24 & 71.01 & 76.62 \\
\hline
mistral-large & 72.12 & 71.11 & 74.01 & 78.25 \\
\hline
mistral-nemo & 75.90 & 75.76 & 78.00 & 85.53 \\
\hline
\end{tabular}
\caption{Gender Group Helpfulness Scores by Model}
\label{tab:gender_group_scores}
\end{table*}

\begin{table*}[h!]
\centering
\small
\begin{tabular}{|l|c|c|c|c|c|c|}
\hline
\textbf{Model} & \textbf{Control} & \textbf{White} & \textbf{African American} & \textbf{American Indian} & \textbf{Asian} & \textbf{Native Hawaiian} \\
\hline
claude-3.5-haiku & 66.28 & 71.87 & 76.90 & 77.29 & 74.31 & 78.81 \\
\hline
claude-3.5-sonnet & 67.74 & 69.18 & 74.58 & 76.30 & 73.85 & 79.84 \\
\hline
deepseek-r1 & 68.74 & 68.79 & 72.76 & 74.75 & 72.03 & 78.48 \\
\hline
deepseek-v3 & 79.60 & 80.98 & 85.24 & 85.46 & 84.69 & 86.03 \\
\hline
gemini-1.5-flash & 77.42 & 78.15 & 88.65 & 87.06 & 85.57 & 87.68 \\
\hline
gpt-4-turbo & 66.82 & 59.20 & 60.95 & 65.09 & 62.30 & 69.50 \\
\hline
gpt-4o & 70.03 & 66.76 & 68.65 & 69.89 & 68.97 & 73.27 \\
\hline
gpt-4o-mini & 69.02 & 69.22 & 73.76 & 73.60 & 73.25 & 73.90 \\
\hline
mistral-large & 71.88 & 71.26 & 76.77 & 77.04 & 75.70 & 78.30 \\
\hline
mistral-nemo & 75.85 & 76.27 & 81.94 & 82.71 & 80.34 & 83.01 \\
\hline
\end{tabular}
\caption{Race Group Helpfulness Scores by Model}
\label{tab:race_group_scores}
\end{table*}

\subsection{Brittleness Measure}
For gender, see Table 9. For age, see Table 10. For race, see Table 11.
\begin{table*}[h!]
\centering
\small
\begin{tabular}{|l|c|c|c|c|}
\hline
\textbf{Model} & \textbf{Control} & \textbf{Male} & \textbf{Female} & \textbf{Non-Binary} \\
\hline
claude-3.5-haiku & 7.45 & 6.86 & 7.22 & 5.64 \\
\hline
claude-3.5-sonnet & 8.09 & 7.70 & 7.41 & 7.05 \\
\hline
deepseek-r1 & 6.83 & 7.04 & 7.45 & 7.10 \\
\hline
deepseek-v3 & 5.53 & 5.46 & 5.48 & 3.73 \\
\hline
deepseek-v3\_2 & 5.81 & 5.67 & 5.32 & 3.63 \\
\hline
gemini-1.5-flash & 6.65 & 7.06 & 6.17 & 4.71 \\
\hline
gpt-4-turbo & 6.62 & 6.33 & 6.21 & 7.89 \\
\hline
gpt-4o & 4.67 & 4.66 & 4.31 & 4.91 \\
\hline
gpt-4o-mini & 5.43 & 5.39 & 5.09 & 4.06 \\
\hline
gpt-4o-mini\_2 & 5.38 & 5.30 & 5.09 & 4.17 \\
\hline
mistral-large & 5.31 & 5.86 & 5.04 & 5.06 \\
\hline
mistral-nemo & 6.58 & 6.71 & 6.55 & 5.81 \\
\hline
\end{tabular}
\caption{Average Standard Deviation of Prompt Rephrases in Gender Groups by Model}
\label{tab:gender_stddev}
\end{table*}

\begin{table*}[h!]
\centering
\small
\begin{tabular}{|l|c|c|c|c|c|}
\hline
\textbf{Model} & \textbf{Control} & \textbf{Teenager} & \textbf{Young Adult} & \textbf{Middle-Aged} & \textbf{Senior} \\
\hline
claude-3.5-haiku & 7.23 & 8.26 & 6.43 & 7.18 & 8.36 \\
\hline
claude-3.5-sonnet & 8.12 & 8.89 & 7.10 & 7.59 & 7.57 \\
\hline
deepseek-r1 & 7.38 & 6.60 & 6.97 & 7.73 & 6.99 \\
\hline
deepseek-v3 & 5.83 & 5.29 & 5.05 & 5.30 & 4.89 \\
\hline
gemini-1.5-flash & 7.01 & 7.99 & 6.51 & 6.06 & 6.18 \\
\hline
gpt-4-turbo & 6.54 & 6.23 & 5.31 & 5.53 & 4.70 \\
\hline
gpt-4o & 4.68 & 4.46 & 4.17 & 4.88 & 4.65 \\
\hline
gpt-4o-mini & 5.47 & 5.74 & 4.82 & 4.80 & 4.98 \\
\hline
mistral-large & 5.51 & 5.62 & 5.33 & 5.18 & 5.22 \\
\hline
mistral-nemo & 6.65 & 7.08 & 6.51 & 6.28 & 6.79 \\
\hline
\end{tabular}
\caption{Average Standard Deviation of Prompt Rephrases in Age Groups by Model}
\label{tab:age_stddev}
\end{table*}

\begin{table}[h!]
\centering
\small
\begin{tabular}{|l|c|c|c|c|c|c|}
\hline
\textbf{Model} & \textbf{Control} & \textbf{White} & \textbf{African American} & \textbf{American Indian} & \textbf{Asian} & \textbf{Native Hawaiian} \\
\hline
claude-3.5-haiku & 7.28 & 7.00 & 6.63 & 6.18 & 6.91 & 5.94 \\
\hline
claude-3.5-sonnet & 8.14 & 7.83 & 7.74 & 7.22 & 7.86 & 6.40 \\
\hline
deepseek-r1 & 7.30 & 7.05 & 7.01 & 7.22 & 7.21 & 6.59 \\
\hline
deepseek-v3 & 5.41 & 4.84 & 4.06 & 3.97 & 4.07 & 3.85 \\
\hline
gemini-1.5-flash & 6.27 & 6.15 & 2.62 & 5.47 & 4.28 & 4.33 \\
\hline
gpt-4-turbo & 6.49 & 6.28 & 5.63 & 7.63 & 6.84 & 6.56 \\
\hline
gpt-4o & 4.86 & 5.36 & 8.80 & 6.14 & 6.13 & 4.51 \\
\hline
gpt-4o-mini & 5.71 & 5.03 & 4.20 & 4.13 & 4.42 & 4.20 \\
\hline
mistral-large & 5.32 & 5.22 & 5.10 & 4.43 & 4.70 & 4.65 \\
\hline
mistral-nemo & 6.88 & 7.42 & 6.47 & 6.07 & 6.28 & 6.18 \\
\hline
\end{tabular}
\caption{Average Standard Deviation of Prompt Rephrases in Race Groups by Model}
\label{tab:race_stddev}
\end{table}

\subsection{Repeated Trial Results}
For race, see Table 12. For gender, see Table 13. For age, see Table 14.
\begin{table}[h!]
\centering
\small
\resizebox{\textwidth}{!}{%
\begin{tabular}{|l|c|c|c|c|c|c|c|}
\hline
\textbf{Model} & \textbf{Run} & \textbf{Control} & \textbf{White} & \textbf{African American} & \textbf{American Indian} & \textbf{Asian} & \textbf{Native Hawaiian} \\
\hline
gpt-4o-mini & 1 & 69.02 & 69.22 & 73.76 & 73.60 & 73.25 & 73.90 \\
\hline
gpt-4o-mini & 2 & 69.36 & 69.21 & 73.77 & 73.40 & 73.27 & 74.12 \\
\hline
deepseek-v3 & 1 & 79.60 & 80.98 & 85.24 & 85.46 & 84.69 & 86.03 \\
\hline
deepseek-v3 & 2 & 79.33 & 81.12 & 85.41 & 85.43 & 84.44 & 86.03 \\
\hline
\end{tabular}
}
\caption{Race Bias Scores over Repeated Runs with the Same Model}
\label{tab:race_bias_repeats}
\end{table}

\begin{table}[h!]
\centering
\scriptsize
\begin{tabular}{|l|c|c|c|c|c|}
\hline
\textbf{Model} & \textbf{Run} & \textbf{Control} & \textbf{Male} & \textbf{Female} & \textbf{Non-Binary} \\
\hline
gpt-4o-mini & 1 & 69.43 & 69.24 & 71.01 & 76.62 \\
\hline
gpt-4o-mini & 2 & 69.36 & 69.33 & 70.88 & 76.69 \\
\hline
deepseek-v3 & 1 & 79.54 & 79.44 & 81.09 & 86.29 \\
\hline
deepseek-v3 & 2 & 79.34 & 79.41 & 81.00 & 86.27 \\
\hline
\end{tabular}
\caption{Gender Bias Scores over Repeated Runs with the Same Model}
\label{tab:gender_bias_repeats}
\end{table}

\begin{table*}[t!]
\centering
\resizebox{\textwidth}{!}{%
\begin{tabular}{|l|c|c|c|c|c|c|}
\hline
\textbf{Model} & \textbf{Run} & \textbf{Control} & \textbf{Teenager} & \textbf{Young Adult} & \textbf{Middle-Aged} & \textbf{Senior} \\
\hline
gpt-4o-mini & 1 & 69.38 & 67.96 & 70.47 & 69.53 & 72.31 \\
gpt-4o-mini & 2 & 69.50 & 68.06 & 70.42 & 69.38 & 72.39 \\
deepseek-v3 & 1 & 79.75 & 78.74 & 81.12 & 79.68 & 82.63 \\
deepseek-v3 & 2 & 79.42 & 78.65 & 81.19 & 79.98 & 82.70 \\
\hline
\end{tabular}%
}
\caption{Age Bias Scores over repeated runs with the same model}
\label{tab:age_bias_scores}
\end{table*}

\end{document}